# Evaluating the Performance of LLMs on Technical Language Processing tasks

Andrew Kernycky, David Coleman, Christopher Spence, and Udayan Das[0000-0003-4393-2644]

Saint Mary's College of California, Moraga, CA 94556
aik1, dlc17, cas36, udd1@stmarys-ca.edu

**Abstract.** In this paper we present the results of an evaluation study of the performance of LLMs on Technical Language Processing tasks. Humans are often confronted with tasks in which they have to gather information from disparate sources and require making sense of large bodies of text. These tasks can be significantly complex for humans and often require deep study including rereading portions of a text. Towards simplifying the task of gathering information we evaluated LLMs with chat interfaces for their ability to provide answers to standard questions that a human can be expected to answer based on their reading of a body of text. The body of text under study is Title 47 of the United States Code of Federal Regulations (CFR) which describes regulations for commercial telecommunications as governed by the Federal Communications Commission (FCC). This has been a body of text of interest because our larger research concerns the issue of making sense of information related to Wireless Spectrum Governance and usage in an automated manner to support Dynamic Spectrum Access. The information concerning this wireless spectrum domain is found in many disparate sources, with Title 47 of the CFR being just one of many.

Using a range of LLMs and providing the required CFR text as context we were able to quantify the performance of those LLMs on the specific task of answering the questions below.

## 1 Introduction

Ever since the introduction of the Transformers architecture[1], language processing has advanced rapidly particularly in the realm of Generative Large Language Models (LLMs). The release of ChatGPT based on the GPT 3.5 LLM in late 2022 [2] has brought this technology to the public domain and there is a great deal of interest in utilizing LLMs and chat tools based on LLMs towards various search and question answering tasks.

That said, systematic evaluation on LLMs and GPT tools on technical tasks are somewhat few, and therefore we were interested in evaluating the performance of several LLM and GPT tools including ChatGPT on a specific subject domain. Our intuition is that generative approaches are by definition limited to the extent that they can provide reliable and citable answers to questions; whereas the subject domain that we are most interested in utilizing automated approaches towards is one that requires a high degree of reliability and, ideally, sub-document level, section-, sub-section, level citability. (We are using the term *citability* throughout this paper as the ability of a model or tool to pinpoint the source of the information contained in a given answer/response.)

Our specific subject domain is technical and regulatory text and information in Wireless Spectrum design, usage, utilization, and governance. The documents and data involved in this domain comes in many forms (discussed in detail in the next section) and involves documents of different types, including, but not limited to, regulatory documents, comments, notices, standards (ex: IEEE 802.11 series [3]), engineering manuals, and license documentation and databases. This is a complex information domain that has been difficult for experienced and knowledgeable humans to parse [4]. Wireless spectrum being a public resource here in the United States and generally the rest of the world, it is essential that all this information be understood more efficiently particularly by those that do not have unlimited resources or decades of experience, including, students, early-career researchers, small and medium-sized and start-up telecommunications companies, and most importantly, policy makers and citizens. The current state-of-the-art as far as language processing does not do an adequate job of working through this complex information domain. Our overall architecture for reasoning depends on the extensive usage of information modeling via knowledge graphs [4–8], however, through several years of continued study we have not found reasonably efficient and reliable Knowledge Graph (KG) creation tools. While some researchers such as the GraphRAG [9] team use LLM-processed Knowledge Graphs (KGs) as a backing for a retrieval augmented generation approach, note here too the quality of the knowledge graphs has not been evaluated and in our experience the quality of knowledge graphs produced are far substandard to painstakingly created human intensive efforts.

This paper reports on the evaluation of a group of LLMs that were being evaluated for their ability to create Knowledge Graphs for use in the Reasoning System architecture. Our intuition was that the tools would be limited but that human augmentation could help us develop a ML-workflow in which the LLMs could play a role. During the course of undergraduate who began by evaluating the LLM/GPT tools on an initial set of *easy* questions (as determined by Prof. Das; the questions are presented later in the paper).

Despite our intuition we were surprised at how middling the responses were and therefore for the purposes of the undergraduate research we put the automated-KG-creation project on pause and undertook a systematic evaluation of the quality of the LLM/GPT tool responses on that initial set of questions. 3 groups of evaluators ranging from novice to expert were chosen to additionally study the impact of LLM/GPT responses on evaluators, particularly around the perceived correctness of the convincing responses. We were surprised on this front that novice evaluators still did a wonderful job of stating that answers were inadequate to their understanding despite perceived confidence of the responses. Novice evaluation is a critical element of our overall approach since the ultimate users of your Reasoning System will include with very limited knowledge of Wireless Spectrum and technologies. Thus it is encouraging that even they bring a healthy level of skepticism when reading responses and also evaluate answers through a lens of understanding and not simply receiving a convincing response.

That may also have implications of LLM/GPT tool usage overall and where they fit in in our overall societal information landscape.

Our current direction is KG-augmented/backed LLMs for Technical Language Processing similar to graphRAG{Citation}, but as stated before we are still in front of high brick wall as far as automating the creation of reliable KGs that can have sub-document level citation information embedded. This is a hard problem.

## 2 A Hard Problem

How is wireless spectrum allocated, how is spectrum managed, and how is spectrum used? These are all complex topics that require understanding complex inter-related information. Spectrum is a limited resource. It is usually considered a public resource in most jurisdictions and there are entities charged with the management of that spectrum. In the United States, the authority that regulates and manages the use of spectrum as a public resource is the Federal Communications Commission (FCC) [10] in conjunction with the National Telecommunications and Information Administration [11]. The former manages spectrum for public use and sets rules for all spectrum usage, and the latter is in charge or managing and regulating spectrum for government applications. The work presented here is focused on the spectrum managed by the FCC.

In the United States, frequency from 9 KHz to 275 GHz is currently allocated. The US frequency allocations [12–14] (see chart shown in figure 1) clearly demonstrates both that spectrum is a limited resource—seen by the extensive allocations, often with more than one application on a given frequency range—and that spectrum allocation and regulation is a complex exercise. The Frequency Allocation Chart shown in figure 1.1 shows which radio services are allocated to various frequency bands and indicates whether each band is used by the federal government or not. Bands are labelled as exclusively for federal government use, exclusively for non-federal government use or shared federal and non-federal government use. Within the bands that are allocated for non-government, public use, the FCC manages the authorization of individual users or operators to use a specific portion of the spectrum at a given place. These authorizations are formalized through FCC licenses. Similar processes govern spectrum regulation outside of the United States as well. There are various types of information that are relevant to the understanding of spectrum:

**Band plans:** the regulatory body defines which portions or bands of the radio frequency range will be used for what purpose. Within bands the body also defines specific information such as channel widths, channel pairings, guard bands, etc. These are published as formal band plans.

**Rules and regulations governing use:** there are specifications set for how communications should occur in a given band. Specifications often define operational parameters such as maximum power output, directionality, distance ranges, maximum heights of antennas, etc. These are usually published in rules and regulations documents. In the United States, these rules can be found in Title 47 of the Code of Federal Regulations (CFR) [15].

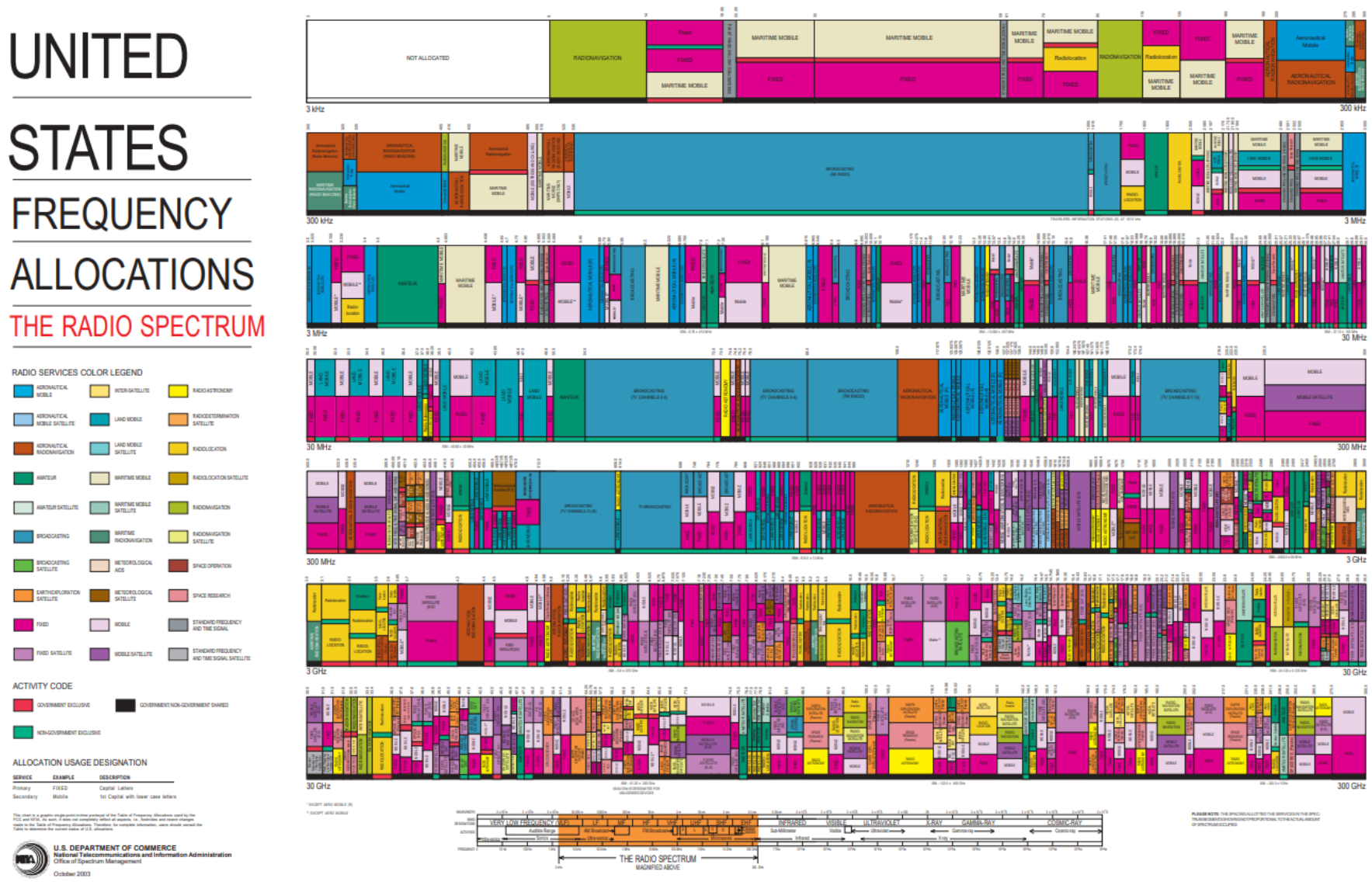

**Fig. 1. United States Frequency Allocation Chart** [13]

**Licensing information and license databases:** the regulatory body allocates channels or frequencies to individual users or operators. The user/operator is authorized to use a certain channel at a given location with specific operational parameters. The regulatory authority typically maintains information on what licenses have been allocated and to whom they have been allocated. [4]presents an extensive discussion of licensing information maintained by the FCC and the modeling of the FCC License database as a KG is discussed in [5]. Understanding wireless spectrum allocation is a considerable challenge due to the complexity and volume of information.

**Spectrum observations:** real-world observations of spectrum usage. The Illinois Institute of Technology Spectrum Observatory project [16] is an example of a long running spectrum observation project. There are many other examples of real-world observations collected as part of various research studies.

**Technical documents:** Many wireless technologies are governed by standards (ex: IEEE 802.11 series of standards [3] that govern Wi-Fi Wireless LAN technologies) and technical/engineering manuals and datasheets may describe their appropriate use, design, characteristics etc.

**Other information:** Businesses that use telecommunications may have predefined hours of operations, these can provide further insight on spectrum usage in a given geo-temporal context. There may also be seasonal patterns to spectrum usage. There are leasing arrangements made between businesses which impact the interpretation of spectrum information. These are just a few examples of other types of information that may be available about specific frequencies including websites and blogs that explain how to interpret specific codes such as Radio Frequency (RF) emission codes.

Each of the above sources of information resides in separate systems, and therefore the information that exists is fragmented across multiple sources (ranging from well-structured to completely unstructured) and there has been very little work to try and connect the different pieces of relevant information. Our system architecture attempts to present the complex inter-related information in a cohesive format so that those interested in reasoning about spectrum can do so.

Each of the sources of information looks different ranging from natural language to decidedly unnatural language (ex: any reader who has never read the IEEE 802.11 standard is encouraged to look at a recent standard like 802.11ac or 802.11ax is challenged to make sense of it). Some sources of information are highly structured (databases) to greatly unstructured (web explainers). So far, it has required (considerable) human intelligence to make sense of the information. There are other domains with similar complexity of information such as healthcare, as well as many other regulatory domains. Various machine approaches though commendable are inadequate in such complex information environments. This is a hard problem.

## 3 The Tools

Table 1 lists all the tools used in this evaluation. We also have data for other model sizes for some of the tools (Ex: H20 GPT was also tested with Falcon 7B and 40 B and Vicuna 33B). Context was provided as Part 101 of the Title 47 of the Code of Federal Regulations (47 CFR part 101) [17]. Being in the public domain and freely on the web, the Code of Federal Regulations (CFR) was found to be part of the training set of all >=GPT3 models. (Context ingestion methods vary from token ingestion to the ability to upload/include a document.)

**Table 1.** LLM/GPT tools evaluated

| Model Name | Model Size (parameters) | Notes |
|---|---|---|
| GPT4All v1.3 Groovy [18, 19] | 6 Billion | Using GPTJ [20, 21] |
| MPT-Instruct [22] | 7 Billion | Based on MPT-7B Mosaic |
| Snoozy-LLM [23] | 13 Billion | Based on LLAMA-13B [24, 25] |
| Vicuna GPT [26] | 13 Billion | |
| H20GPT [27, 28] | 65 Billion | Using GPT3.5 Turbo [29] |
| ChatGPT [2] | Estimated at 170 Billion+ | |

## 4 The Questions and Answer Evaluation

The following questions were used for the evaluation. These are questions that merely producing word statistics are unlikely to answer statistically. However, to be useful tools LLMs should be evaluated for tasks that are complicated. This task is in our estimation a “modest” difficulty task for someone starting in telecommunications (student, researcher, or new enterprise). The overall situation as shared in the previous section is an “extreme” difficulty task. We decided to begin our evaluation with this task expectin that we would then increase the difficulty by choosing a cross-reference question evaluation. As reported here the limitations of the responses suggested that we do a detailed evaluation of these questions. (Note that this is work performed by undergraduate researchers as part of a summer project.)

1. What is HAAT?
2. What is LMR?
3. What is the maximum bandwidth authorized per frequency in the following bands: 3700 - 4200 MHz, 18,142 - 18,580 MHz, and 928-929 MHz.
4. Which of the 928 MHz channels are allowed to be 12.5 KHz?
5. What types of bands are 3700 - 4200 MHz, 18,142 - 18,580 MHz, and 928-929 MHz.
6. What is the minimum path length requirement for Fixed Microwave Services at 6 GHz with 1 GHz above and below?
7. What does EIRP mean?
8. What does this mean: maximum bandwidth authorized per frequency
9. Which frequency bands can Fixed Microwave Services potentially interfere with geostationary applications in the fixed satellite service?
10. Which section describes how the Fixed Microwave Services can minimize the probability of interference to geostationary satellites?
11. BONUS: How to find the geostationary orbit so that we can be outside 2 degrees of it (taking into account atmospheric refraction).

As can be seen these questions are extremely domain specific and requires some knowledge of the subject. The answers can be found in Title 47 of the CFR[30, 31]. Most of the answers, outside of the definitions, can be found in part 101[17]. We made the choice to limit questions to a small part of the text so as to be able to evaluate the models without running into token limit issues and to control costs.

### 4.1 Data Collection and Survey Design

The survey was administered using Google Forms as the data collection platform. The survey consisted of a questionnaire designed to gather information on participant comprehension with the answers generated by each LLM. Evaluators were designated into 3 types: 1) novice: people who are unfamiliar with wireless technology and telecommunications; 2) techie: people who were familiar with wireless technology but had limited understanding of telecommunications standards and policies; and 3) expert: an individual who has extensive knowledge of all of the topics at hand. These surveyors had varying degrees of familiarity with the FCC codes and regulations ranging from zero to expert experience.

Depending on their proficiency level, each participant was not expected to fully understand if the information was correct, but rather their perception of comprehensibility and how convincing they found the answer.

The survey included a Likert scale for each question and model with options ranging “1-10” from "No Understanding" to "Easily Understood“. Each question was provided to the surveyors along with a clearly marked which model had generated the response. Additionally, an optional open response section was provided for participants to provide additional comments or insights to further future research and development. Specifically, of interest was a “say more” response from several evaluators for several responses.

### 4.2 Evaluation

A team of human evaluators rated the answers provided. Answers were evaluated for comprehensibility and perceived correctness. The team of evaluators constituted individuals with almost no background in wireless spectrum, individuals with some knowledge of wireless spectrum, and an expert who is well-versed. Any evaluation must include a variety of evaluators because LLMs have the tendency to come up with convincing hallucinations. At the same time, individuals with less background can still be great evaluators because they can quickly anticipate the dictionary problem. The dictionary problem is the problem of looking up a definition and being pressed to look up many other definitions in order to make sense of the original definition. In our study, in many instances human evaluators found that the responses left them with a chain of future questions, and they would have liked to have asked the LLMs follow-up questions. Our objective was not perfect understanding for our evaluators so we did not allow follow-up questions. We also believe that in a question-answering context it is important that the answering entity recognize the comprehension level of their audience and modify answers accordingly. For example, when an instructor answers questions in class, they do not give an abstract answer that will require students separate study to understand; instead, the instructor curates the answer to their audience.

## 5 Results

Figure 2 below shows a summary of the average and median for evaluator scores for all evaluated tools along with an Expert Score (ES) column to demonstrate the divergence between novice/general perception and expert evaluation.

| | GPT4All | | | MPT-InstructLLM | | | Snoozy LLM | | | VicunaGPT | | | H20GPT | | | ChatGPT | | |
|---|---|---|---|---|---|---|---|---|---|---|---|---|---|---|---|---|---|---|
| | Avg | Md | ES | Avg | Md | ES | Avg | Md | ES | Avg | Md | ES | Avg | Md | ES | Avg | Md | ES |
| **Q1** | 5.83 | 6.5 | **1** | 5.33 | 6.5 | **1** | 6.83 | 8.5 | **1** | 6.67 | 7 | **2** | 8.17 | 8 | **7** | 8.33 | 8 | **8** |
| **Q2** | 3.83 | 3.5 | **3** | 3.17 | 3 | **1** | 7.33 | 8.5 | **3** | 6.67 | 7 | **5** | 8.17 | 9 | **4** | 8.5 | 9 | **7** |
| **Q3** | 7 | 7 | **3** | 6.83 | 7.5 | **5** | 5.67 | 5 | **6** | 8 | 8.5 | **8** | 7.67 | 8 | **3** | 7.33 | 8 | **6** |
| **Q4** | 5.5 | 5 | **3** | 3 | 3 | **1** | 5.33 | 4.5 | **3** | 7.16 | 7.5 | **2** | 7 | 8 | **2** | 7 | 6.5 | **5** |
| **Q5** | 6.33 | 7 | **1** | 5.83 | 6.5 | **1** | 5.67 | 5.5 | **5** | 8.16 | 8 | **6** | 8.67 | 9 | **7** | 8.33 | 8 | **10** |
| **Q6** | 2.67 | 1 | **1** | 5.17 | 5 | **1** | 6.33 | 7 | **3** | 6.17 | 7.5 | **1** | 7.33 | 7 | **5** | 8.33 | 8.5 | **7** |
| **Q7** | 5.17 | 5.5 | **1** | 8.33 | 9 | **5** | 8.83 | 9 | **7** | 7.33 | 7 | **5** | 8.67 | 8 | **8** | 9.33 | 10 | **8** |
| **Q8** | 8.67 | 9 | **8** | 5.17 | 5 | **1** | 7.67 | 7.5 | **7** | 7.17 | 6.5 | **6** | 8.83 | 9 | **6** | 9 | 9 | **7** |
| **Q9** | 3.83 | 3.5 | **1** | 7.83 | 8.5 | **8** | 7.67 | 10 | **1** | 7.83 | 8 | **8** | 7.83 | 9 | **9** | 6.5 | 6.5 | **4** |
| **Q10** | 3 | 2.5 | **1** | 5.83 | 5.5 | **3** | 6.17 | 6.5 | **1** | 4.17 | 3 | **1** | 6.67 | 7 | **1** | 6.83 | 6.5 | **6** |
| **Bonus** | 3.83 | 3 | **1** | 3.67 | 3.5 | **2** | 6.5 | 6 | **6** | 6.67 | 7 | **3** | 8 | 8.5 | **6** | 6.83 | 6.5 | **7** |

**Fig. 2. Scores by question and tool** (shows Avg:Average; Md:Median; **ES:Expert Score**)

## 6 Discussion

To call information inaccuracies in LLMs "hallucinations" are a misnomer. By design, these models are not outputting accurate information but convincing text in the language(s) of training. Although LLMs have shown promise in information retrieval (IR) tasks, their information retrieval capabilities are accidental and not by design. The primary purpose of LLMs has been to advance NLP by extracting statistical relationships in language and thereby improve the comprehensibility and quality of responses to human questions. That these models have demonstrated good question-answering capabilities is a testament to the history of the written word in human civilization and how well human beings have structured various forms of knowledge in text. To expect the learning of these models to excel in IR tasks demonstrates a limited understanding of how they work. We will for the purposes of this paper leave out critiques of whether not these models are verging on AGI. The limitations indicated in this paper and discussed in brief detail in this section, should, to the intelligent observer, make the argument for us. In the next section we will discuss future work and approaches towards improving the responses received from an LLM, particularly on critical tasks in which reliable information and *citability* is critical.

Our findings indicate that the answers are often middling at best, and hallucinations are a very real problem. As can be seen in table 2, the performance improves with model size with ChatGPT performing the best of all, however, there is not a consistent improvement in response quality. Note the expert rating (ES column) for each model. Even though ChatGPT scores a 10 in one response, the rest of responses have an average of 5.9 Expert rating which is not very good for the given task.

When evaluating responses, it becomes clear that the participants' experience and expertise played a critical role in their perception of answer quality. A common trend noticed was how the more confidence and higher level of language clarity used would push participants to rate the model higher. However, as the expertise level increases, participants begin to identify inaccuracies and off-topic information even in answers that had these two qualities.

When evaluating Snoozy it was found that it tended to provide answers that were more readily accepted by those with limited knowledge of the subject matter. While those with higher expertise, begin to identify inaccuracies and off-topic information resulting in their ratings to be much lower as they lose confidence in the response.

MPT-Instruct LLM displayed a similar pattern, providing relatively convincing answers to participants with little experience but failing to meet the expectations of experts. In both cases, experts found the answers inadequate or off-topic, highlighting the limitations of these models in providing precise technical information.

With H2OGPT (GPT 3.5 Turbo), the model would often lean towards the right answer, but lacked objectivity and correctness. While the model could stay on topic and near relevant information, there were major tendencies to stray away from specific answers and needed context. Experts often found that H2OGPT's results were generalized, lacked context, and the model's answers were too broad for what was asked.

VicunaGPT's results were always given in direct readable formats, and quite convincingly displayed incorrect information for every question given. The model gave incredible confidence with little-to-no reason behind the answers, and nearly tricked every non-expert surveyor, every time. Compared to the other 13 billion and below parameter models used in this research, VicunaGPT was proven to be exceedingly well spoken and wildly inaccurate.

It was found that OpenAI's ChatGPT model resulted in the highest average scores throughout the questionnaire, while H2O GPT followed close behind. It was noticed that as models had less access to parameters and data their capabilities to provide comprehensible and accurate answers suffered. Participants were quick to note how many of the answers some of these models provided were avoiding the question or providing information that did not seem relevant. Participants' feedback provided additional insights into the strengths and weaknesses of these generated responses. participants often noted the need for more specific information, clear definitions, and context.

Recent work by Schaeffer et al. [32, 33] has questioned the rapid scaling of LLM abilities, suggesting that different evaluators have used different metrics to show advances but using similar metrics the scaling is found to be linear with the scale of the models with an absence of "emergent" capabilities. We do not find the result at all surprising. Significant IR and QA challenges remain and research should be encouraged in improving performance at a given scale with adequate language capability and not unlimited resources towards ever larger scaling.

As Computer Scientists we should also consider the cost of these ever larger models. As researchers we should continue to insist on more open conduct of research and more equitable allocation of resources. The tremendous progress made in language models has been largely driven by both a vast variety of researchers as well as openness around the research conducted. With large companies controlling very large language models, it is unclear if the community at large stands to benefit.

And lest it be forgotten, it is worth reminding everyone that a majority of foundational work in AI (of all kinds) originated in academic research.

## 7 Conclusion

In summary, the data suggests that participants with less experience and knowledge in FCC-related topics are more likely to be more confident in the answers provided by these models, even when these answers are inaccurate or off-topic. As expertise levels increase, users become more critical of the LLM-generated responses, identifying inconsistencies and shortcomings.

These findings highlight the importance of caution when relying on language models for technical information. Users must be aware of the limitations of these models and critically evaluate the responses they receive, especially in domains that demand precision and accuracy, such as the Federal Communication Commission's regulations. Large language models can be a valuable tool, but they should assist human expertise rather than replace it.

Significant research effort should be devoted in making question-answering reliable with sub-document level *citability* for critical technical tasks instead of scaling LLMs to infinity.

**Disclosure of Interests.** This work was funded by the Saint Mary's College School of Science Summer Research Program for Undergraduates.